\documentclass[a4paper,man,natbib, floatsintext]{apa6}

\usepackage[english]{babel}
\usepackage[utf8x]{inputenc}
\usepackage{amsmath}
\usepackage{graphicx}
\usepackage{placeins}
\usepackage{afterpage}
\usepackage[colorinlistoftodos]{todonotes}
\usepackage{xspace}

\title{Implicit Dimension Identification in User-Generated Text with LSTM Networks}
\shorttitle{Implicit Dimension Identification in User-Generated Text}
\author{Victor Makarenkov, Ido Guy, Niva Hazon, Tamar Meisels, Bracha Shapira, Lior Rokach \\
\{makarenk, nivah, tamar, bshapira, liorrk\}@post.bgu.ac.il , idoguy@acm.org}

\affiliation{Department of Software and Information Systems Engineering \\ Ben-Gurion University of the Negev}

\abstract{
 In the process of online storytelling, individual users create and consume highly diverse content that contains a great deal of implicit beliefs and not plainly expressed narrative. It is hard to manually detect these implicit beliefs, intentions and moral foundations of the writers. 
 
We study and investigate two different tasks, each of which reflect the difficulty of detecting an implicit user's knowledge, intent or belief that may be based on writer's moral foundation: 1) political perspective detection in news articles  2) identification of informational vs. conversational questions in community question answering (CQA) archives and. In both tasks we first describe new interesting annotated datasets and make the datasets publicly available. Second, we compare various classification algorithms, and show the differences in their performance on both tasks. Third, in political perspective detection task we utilize a narrative representation language of local press to identify perspective differences between presumably neutral American and British press.
 
}

\begin{document}
\maketitle

\newcommand{\ig}[1]{\textcolor{blue}{$\ll$\textsf{#1 --IG}$\gg$}}
\newcommand{\YA}{{YA}\xspace}
\newcommand{\bs}[1]{\textcolor{red}{$\ll$\textsf{#1 --BS}$\gg$}}
\newcommand{\lr}[1]{\textcolor{blue}{$\ll$\textsf{#1 --LR}$\gg$}}
\newcommand{\vm}[1]{\textcolor{purple}{$\ll$\textsf{#1 --VM}$\gg$}}
\newcommand{\harper}{{\citep{harper09facts}}\xspace}

\newcommand{\lp}{{LP}\xspace}
 \section{Introduction}


Creating new narratives and publishing stories are not limited to profit-oriented publishing organizations of newspapers, printed magazines and books.
The number of users taking advantage of interactive opportunities on the Internet to share their stories continues to increase. The combination of active and creative participants and a similarly engaged audience, along with accessible and easy to use platforms \citep{SHELDON201689}, tools, and applications, has created a new culture in which users express their ideas and opinions and provide information on various domains.  The analysis of user-generated content (UGC) while modeling existing narratives \citep{NarrativePhd} can offer insights into individual and societal concerns and benefit a wide range of applications, for example, tracking mobility in
cities, identifying citizen’s issues, and opinion mining. In contrast to traditional forms of
media and content, UGC varies greatly in terms of length, content, quality, language,
 and other aspects. This heterogeneity poses new challenges to text analysis methods.

In a recent report \citep{dugshtal} from the Dagstuhl Seminar about user generated content in Social media the authors defined nine measurable text properties. These properties form the "Information Nutrition Label" inspired by the nutritional labels found on most food products.  These measures indicate the “objective” value of User Generated text and include: factuality, reading level, virality, emotion, opinion,
controversy, authority, technicality, and topicality. 
For example, the \textit{opinion} property of a text reflects the percentage of "opinionated" sentences, while opinion mining is the task of performing sentiment analysis on these opinionated sentences to learn the opinion (negative or positive) of the user towards some entity, as explicitly stated in the opinionated sentences.  
The  \textit{emotion} property is a measure of how emotive is the text; i.e.,  how many emotional arguments that employ words charged with positive or
negative connotations are included in the text.
The \textit {controversy} property measures the level of contradiction in the points of view that some texts reflect.
The report summarizes the state of the art machine learning methods that can be used to mitigate the challenges associated with assessing the above mentioned properties in UGC.

While these properties are general and relevant for a well defined factual narrative, storytelling may include one's own beliefs, which can be either disputable or agreed on. User stories often have  implicit subjective dimensions which are relevant in the context of a specific domain and reflect the author's narrative rather than the language used. For example, academic papers presenting scientific studies can be characterized as  describing technical, theoretical, or applied science; or a short story can be characterized by the age group that the author was targeting. For such cases it would be useful to identify and classify implicit properties that typically don't have well-defined explicit indicators but are clearly implied from the text when manually annotated.  In this paper we propose a method for handling implicit dimension identification in text and demonstrate and evaluate it on two relevant use cases. 


The told and intended narrative can be \citep{NarrativePhd, chambers2009unsupervised} either factually right or someone's own belief. We study two problems in order to analyze the ability to automatically detect and identify these user's intended narratives.

The first use case is based on political perspective identification in online news articles. This is important for assessing the subjectivity or neutrality of news reports and sources. We classify the reported article as either neutral or leaning towards one actor involved in a political conflict. Like in many different political conflicts in different countries, the Israeli-Arab conflict also has no one single truth everyone agrees on. We exploit Israeli and Palestinian media to show the detection of each side's narrative in presumably neutral American and British press.

The second use case we present is based on the task of classifying informational versus conversational questions, which is important and relevant in the context of community question answering (CQA). Questions on CQA websites usually reflect one of two intents: learning information or starting a conversation. Many of the uses of CQA archives, such as supporting question retrieval or serving CQA-intent queries on Web search  suit one of the two question types (Harper et al., 2009) and can therefore benefit from automated classification of these implicit dimensions.  

Both cases reflect tasks that have latent implied properties that are not directly expressed in the text and are specific to their domains.

Implicit dimension identification is a complicated task, and implied properties are not always easily determined, even by humans, and may require thorough reading of the text. For example, a community question like \textit{"Who is ultimately responsible for the oil spill in the gulf coast?"} is probably more conversational than informational as there is no ultimate answer, and the question would likely stimulate discussion. 
However, the decision whether this and other questions are conversational or informational is not very clear, and humans need to discuss and define the boundaries between the classes.

We suggest using an RNN with LSTM networks to identify the implicit dimensions in a piece of text for each of the above mentioned tasks and demonstrate its performance on both tasks.
As opposed to many standard machine learning classifiers including feed-forward neural networks, RNNs impose an order on the components of every instance in the data during the training and prediction \citep{Goodfellow-et-al-2016}. Because the \textit{order of a sequence} is a natural property of the words within a piece of text, 
new training paradigms were suggested in recent years. The boundaries between supervised and unsupervised methods became blurred; for example, in the RNN language models each word can first serve as an instance and immediately after serve as the label for the prefix \citep{DBLP:conf/slt/MikolovZ12}. Moreover, adding an LSTM cell to an RNN makes its training procedure easier
\citep{Hochreiter:1997:LSM:1246443.1246450} and the resulting classifier's performance more robust. LSTM allows complex linguistic regularities to be captured; such regularities are harder to exploit using a standard feature \textit{set} without ordering. Traditional word statistical features like n-grams usually take into account  unigram, bi-gram, and tri-gram, and less frequently four or five-gram. Although these standard features of textual data perform well in many tasks, we show that longer, multiple sentence and paragraph cutting features further improve implicit dimension identification performance. LSTM networks impose sophisticated weighted ordering on the sequence of words within a piece of text during the training and classification stages. 

The following points summarize the contribution of our paper:
\begin{itemize}
\item We study two cases of implicit dimensions identification, where there is no precise agreement on the classified subjects inside the intended narrative. We support the two use cases with two interesting datasets, experiment with various classification algorithms to analyze the intended narrative.We make the datasets publicly available.
\item We show how unlabeled data, which is typically more available in classification tasks than labeled data, can be effectively utilized to enhance performance in the text classification domain. We show how unlabeled pre-trained word embeddings can be used to improve classification accuracy combined with LSTM models. 

\end{itemize}

\section{Background and Related Work} 

It is hard to detect implicit narrative in UGC due to complex language properties such as irony, lack of capitalization, spelling errors, varying length, and heterogeneous quality \citep{Sarmento:2009:ACR:1651461.1651468, DBLP:journals/dagstuhl-reports/ChuaFGJP17} 
due to the fact that anyone can post text online with very little content moderation.
We follow common practice for this task and utilize text classification techniques. We focus on the RNN classifier to address the two previously described tasks. 
\label{section:rnn}

\subsection{Scientific Background}

The folklore feed forward neural network, also known as the multilayer perceptron (MLP) has been heavily used as a classification technique in classification applications \citep{Guy:2018:IIV:3159652.3159733,harper09facts}

The more recently popularized LSTM \citep{Hochreiter:1997:LSM:1246443.1246450} network is a robust, widely-used, and high-performing classifier. It can be trained easily due to its complex structure and ability to handle vanishing gradient phenomena.  

LSTM networks have been used successfully in several NLP applications \citep{baccouche2010action,graves2005framewise, DBLP:journals/corr/BadjatiyaG0V17, DBLP:journals/corr/WuSCLNMKCGMKSJL16, sutskever2014sequence, DBLP:conf/acl/AharoniG17,melamud2016context2vec, P16-2067, DBLP:journals/corr/RuchanskySL17}.

LSTM networks have been heavily adopted because they can capture very long linguistic regularities and complex patterns within a given text.

In addition to the adoption of LSTM networks for the whole sequence classification, it has also been used as a sequence labeling mechanism, for example, in the part-of-speech (POS) labeling task \citep{graves2012supervised, DBLP:journals/corr/WangQSHZ15}; in this task a classification should be made in each step $t$ of the network unrolling. The language modeling task \citep{zaremba2014recurrent, jozefowicz2016exploring, shazeer2017outrageously, DBLP:journals/corr/MerityXBS16} represents another use of RNN and LSTM for the underlying structure of sequence classification; this task recently gained some attention when LSTM methods were introduced. 

LSTM has been applied in various text classification tasks, such as sentiment analysis \citep{ISI:000379634700032, ISI:000392770900019, ISI:000397687400019, ISI:000414619600010}, currently a hot topic of fake news detection \citep{Ma:2016:DRM:3061053.3061153,DBLP:journals/corr/abs-1708-01967,DBLP:journals/corr/RuchanskySL17} and
hate speech detection on social networks \citep{DBLP:journals/corr/BadjatiyaG0V17}.

\section{The Task of Implicit Dimensions Identification}
\subsection{Identifying a Political Perspective in Online News Articles}

News subjectivity is a phenomenon that has generated a wide range of research studies in the academic worlds of communications, media and computer sciences. Many of the media investigations were performed using expensive manual content comparison and analysis. Studies of subjectivity in the computational field have concentrated on detecting patterns automatically with the goal of automating the analysis of large amounts of data. 
Personal perspective can potentially be present in every news report. There are many ways for personal perspective to present itself, i.e., stating personal opinions, providing incorrect facts or figures, applying unequal space to different sides of a controversial issue and interviewing more people on one side of the controversy. This behavior can extend beyond individual posters to editors as well, and their role may also include deciding which news stories to cover and publish, and what stories to present on the front page.

\textit{Bias} in cognitive science is generally defined as a “deviation from some norm or objective value” \citep{mladenic2015learning}. Every news story has a potential to be influenced by the personal opinion and background of the journalist or news source and therefore potentially biased. 

News bias can be reflected in many ways – in any given story some details can be ignored and others included, and how the story is organized and framed. Headlines can contain bias by expressing approval or condemnation. Word choice and connotation is extremely important as well. Images- choice of photo, the angle and the caption below the image can be significant. Bias via the description of   people places and events, for example “terrorist” or “freedom fighter” can be crucial.  In covering controversial issues one can apply unequal space to different sides and cite more representatives on one side of the conflict.  More severe bias can include incorrect facts or figures.

Because the term bias is very emotionally charged, in the context of national and disputed issues we investigate the \textit{perspective} presence in the news.

\subsubsection{Related Work on Bias Detection in the News}

\citep{fortuna2008detecting} detect subtle biases in the content of four online media outlets: CNN, Al Jazeera (AJ), International Herald Tribune (IHT), and Detroit News (DN).  This work focuses on two types of bias: which stories to cover, and which terms to use when reporting on a given story. It further shows that algorithms from statistical learning theory (specifically kernel-based methods) can be combined with ideas from traditional statistics, in order to detect bias in the content of these news sources. Their automated analysis has uncovered the existence of a statistically significant lexical difference between CNN and Al Jazeera in the way they report the same events.

  \citep{groseclose2005measure} also measure media bias (“slant”) by estimating and grading ideological scores for several American media outlets. They counted the number of times a particular media outlet cites various policy groups and think tanks listed on the website and then compare it with the number of times members of the United States Congress cite the same groups or politicians. Their results revealed a strong liberal bias in most of the online outlets under observations.

\citep{park2009newscube} developed and evaluated the NewsCube system which aims to mitigate the effect of media bias. NewsCube classifies content into different viewpoints or “aspects” on issues. It supports “aspect-level browsing” by providing readers with sets of articles with different aspects and groups together a set of articles by the similarity of their aspects. Classification of aspects is done with an unsupervised classification method- clustering. However, supervised classification is not possible since they cannot predict and construct a pre-defined category of aspects for a news event. The results showed that this ‘different viewpoint’ presentation made the subject want to read more articles and compare different articles on topic, even opinions they disagreed with.

A recent project \citep{sholar2016predicting} uses Event Registry \citep{leban2016event} API to collect news with the goal of analyzing media bias.  It focuses on Israel and Palestinian news.  The first part of this project uses Naïve Bayes and SVM models in order to predict the news source, given the headline. The second part is a study of event selection bias by examining three models for each news source- number of articles covering each event, binary indicator if event was covered, and normalized propensity. Each model is then used to cluster the news sources.  It seems that the Event Registry does show differences among news sources but does not indicate any specific biases.

It appears that no previous work has been done using thousands of articles from major media outlets to automatically detect perspective using machine learning models, and specifically covering the Israeli-Palestinian conflict. Furthermore, no previous research uses the method of analyzing articles from Israeli and Arab news sources to help detect bias and stances in international news articles.

\subsubsection{Perspective Identification in the Israeli-Palestinian Conflict}

In this task we use machine learning techniques to detect narrative, perspective, and leanings in news articles and their sources which are considered neutral. Specifically, we examine the idea in the context of the Israeli-Palestinian conflict. Comparison among news sources and news articles will inform news consumers of the leanings of the news source they are consuming. We handle the problem of identifying such perspectives as a classification task.

The Internet has emerged as the primary source of information for many Americans \citep{doi:10.1177/1461444809350193} and others. Therefore, analyzing online news and detecting bias and variance among online news sources is an important task.

More specifically, the Israeli-Palestinian conflict is one of the most debated and frequently covered issues in European and American news. Media coverage of the Arab–Israeli conflict by journalists in international news media has been said to be biased on both sides.

For the purpose of this paper, we are generally referring to the coverage of Palestinian-Israeli conflict. Certainly, there is much overlap with the greater Arab-Israeli conflict, but we will try to be more specific and limit it to the coverage of Palestinian-Israeli disputes and issues.  Naturally, 'Arab media' refers to the larger region and not only Palestinian media.
We also check the leanings of European and American press.
We refer to this task as the \textit{NEWS task} in this work.

\subsection{Identifying Informational vs Conversational Question on CQA Archives}
Questions on CQA websites span many domains and involve a variety of user needs, which generally map into two main types, defined by \citep{harper09facts}: 1) \textit{Informational} questions which are asked with the intent of obtaining information that the asker hopes to learn or use via fact- or advice-oriented answers, and 2) \textit{Conversational} questions which are asked with the intent of stimulating discussion and may be aimed at gathering opinions or reflect an act of self-expression \citep{harper09facts}. While their paper focused on classifying questions at posting time, in order to improve question routing and automated tagging in real time, we consider CQA archives~\citep{xue08retrieval}, which allow us to examine additional information that accumulates after the question has been posted, such as answers and votes. Many of the uses of CQA archives suit one of the two types of questions and can therefore benefit from automated classification. Supporting question retrieval~\citep{jeon20finding} and serving CQA-intent queries on Web search~\citep{levi18selective}, perhaps the two most common practices of CQA archives, correspond to the informational type questions, and with the recent advancements in social media mining, conversational questions in the archives can be used to support a variety of applications, such as opinion mining~\citep{pang08opinion}, controversy detection~\citep{dori15automated}, and automated debating~\citep{gurevych16debating}.

\subsubsection{Related Work on CQA Analysis}

Within CQA research, our work falls under the broad category of content modeling~\citep{srba16survey}, which includes areas such as question quality estimation (e.g.,~\citep{li12analyzing}), answer quality ranking (e.g.,~\citep{toba14discovering}), question topic classification~\citep{cai11large
}, and question type classification~\citep{liu09urgent}. Some of the CQA research along the years has focused on specific types of questions such as factoids (e.g.,~\citep{bian08factoid,guy16factoid}), advice-seeking~\citep{braunstain2016supporting}, how-to questions ~\citep{surdeanu11learning,weber12answers}, why questions~\citep{oh12why}, and opinion questions~\citep{liu08evolution}. Some of these types have stronger association with the informational class (e.g., factoids, how-to), while others are more connected to the conversational class (e.g., opinion, why). 
A good summary of CQA research can be found in the recent survey by Srba and Bielikova~\citep{srba16survey}.

More specifically, our work focuses on user intent classification on CQA. User intent has been extensively studied in Web information access, most prominently in Web search. The seminal work by Broder~\citep{broder02taxonomy} distinguished between three main types of Web search queries: navigational, informational, and transactional. At a high level, all three map to the informational class in CQA, as Web search does not involve any aspect of conversation among users. Later works refined Broder's taxonomy and developed automatic classifiers to distinguish between the types (e.g., ~\citep{jansen07determining,brenes09survey,zha10visual}). On the other hand, on social media, user intent typically revolves around sharing, interacting, conversing, and socializing~\citep{java07why,lin11people,ellison07benefits}. To a certain extent, CQA websites combine both of these worlds, as they involve explicit user input in the form of a question, but also enable some level of user interaction. 
%
%


Our work is based on the definitions of informational and conversational questions by \harper. 
Several studies examined user intent on question answering systems from other angles. Prominently, the distinction between objective and subjective questions was explored in several works~\citep{li08cocqa,chen12understanding,aikawa11community}. While this distinction bears some similarity to the informational and conversational division, it is not the same:  
the subjective class includes, by definition~\citep{chen12understanding}, general advice, which is normally considered informational by Harper's definitions. For example, the question \textit{``I am a Bangladeshi National girl and I came to USA on B1/B2 visa and now I would like to take admission pls adv?''}, given as an example for a subjective intent~\citep{chen12understanding}, is informational. Indeed, as reported over a \YA dataset~\citep{li08cocqa}, only 34\% of the questions were labeled as objective, while in \harper and our work, 61.2\% and 55.6\%, respectively, were labeled informational.


Mendes Rodrigues and Milic-Frayling~\citep{mr09:socializing} experimented with both \YA and MSN QnA, a CQA website that was closed in 2009. 
They defined ``social'' intent for questions that are posted in order to informally engage and interact with other community users, as typically occurs in chatrooms. Social intent implies conversational intent, but the latter is  far more comprehensive, also reflecting other needs, such as polling or discussing a topic, without socializing. Indeed, social intent was more common on MSN QnA, which enabled more intense user interaction through flexible comments and tagging, rendering a rich thread structure. A classifier deemed 6.5\% of the questions on MSN QnA as social, while on \YA this type's occurrence was ``not sufficient to train a classifier''~\citep{mr09:socializing}.

\subsubsection{The CQA task in Yahoo! Answers}
As our CQA website, we opted to focus on Yahoo! Answers (\YA), one of the largest CQA websites. In 2008, it was reported to account for 74\% of CQA traffic~\citep{harper09facts}, and while somewhat decreasing in popularity since then, it still enjoys tens of thousands of questions posted every day, allowing us to inspect a decade's worth of queries. Aside from examining a long time period, we also experiment with a substantially greater amount of data: the proprietary dataset of Harper et al.'s included only 151 labeled questions from \YA, while we build a dataset of over $4000$ labeled questions, which are publicly available. 
Additionally, we examine the use of millions of unlabeled \YA questions to further enhance performance. We believe that the larger dataset, as well as the use of additional textual fields and the advancement in machine learning methods, 
all warrant revisiting the informational/conversational classification task. In the remainder of this work, we refer to this task as the \textit{CQA task}.



\section{Datasets}
\label{sec:datasets}
\subsection{Online News Articles Dataset}

\subsubsection{Gathering news articles}

We gathered online news articles discussing the Israeli-Palestinian conflict from Israeli and Palestinian news sources as well as international media outlets from the years 2014 to 2016.
We used the \url{eventregistry.org} API. Event Registry collects news articles from RSS feeds of over 100,000 news sources around the world. We collected 25,000 news articles regarding the Palestinian-Israeli conflict from these sources. Table \ref{table:news:source} presents the annual statistics per year for the collected articles.  

\begin{table}[]
\centering
\caption{News articles collected from several news media outlets. Summarizing the Israeli, the Arab and the international news sources statistics.}
\label{table:news:source}
\begin{tabular}{lcccc}
\hline
 & \textbf{\begin{tabular}[c]{@{}c@{}}Total number\\ of articles\end{tabular}} & \textbf{2014} & \textbf{2015} & \textbf{2016} \\ \hline
Jerusalem Post & 6022 & 3159 & 1641 & 1222 \\
Arutz Sheva & 4025 & 2830 & 550 & 645 \\
Times of Israel & 1780 & 147 & 61 & 1572 \\
Hamodia & 739 & - & 196 & 543 \\ \hline
English.Pal info & 2328 & - & 431 & 1897 \\
Wafa & 610 & 125 & 485 & - \\
Palestine Chronicle & 1133 & 113 & 532 & 488 \\
Al-Jazeera (Qatar) & 1069 & 310 & 350 & 409 \\
Al-Jazeera (Saudi) & 851 & 450 & 244 & 157 \\
Al-Bawaba (Jordan) & 1640 & 520 & 512 & 608 \\ \hline
BBC & 536 & 163 & 201 & 90 \\
The Guardian & 269 & 158 & 92 & 74 \\
Fox News & 1299 & 416 & 512 & 371 \\
The New York Times & 1980 & 785 & 819 & 376
\end{tabular}
\end{table}

\subsubsection{Article composition}

In this work we considered only two signals from each article: the article's \textit{title} and its \textit{content}. We didn't consider its published date, source, or the author's name. No external data was used in the classification; we used only the text of the article's title, content, or their concatenation.

\subsubsection{Labeling methods}

For training part of building the classifiers, we used Israeli and Arab news sources; the articles from the Israeli news sources are labeled “Israeli perspective,” while the articles from the Arab news sources are labeled “Palestinian perspective.”
\begin{figure}[h!]
    \centering
    \includegraphics[width=0.7\textwidth]{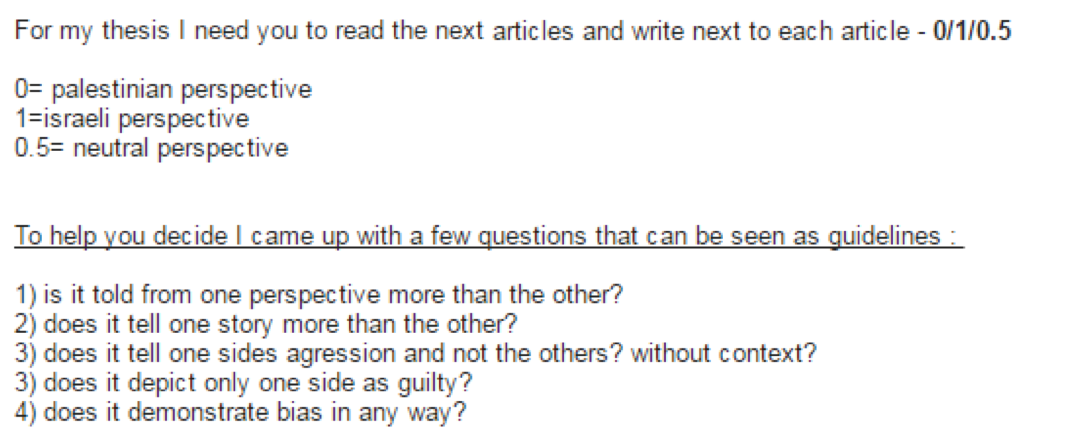}
    \caption{The instructions that were given to the annotators.} 
    \label{fig:instructions}
\end{figure}
\FloatBarrier
For the evaluation part we used human annotators who classified international news articles as either Israeli perspective, Palestinian perspective, or neutral\footnote{The labeled data available at \url{https://github.com/vicmak/News-Bias-Detection/}}. 

For instance the following BBC-news article was agreed by the annotators to have a Palestinian perspective: "\textit{Palestinian flag to be raised at United Nations}". A Fox-news article titled "\textit{Wave of rocket attacks on Israel signal power struggle in Gaza}" was jointly annotated as having an Israeli perspective.

\subsubsection{Human annotators}
159 articles which were not used in the training set were chosen from the original set of international articles. Two human annotators were employed to assess the perspective of the article. This was done in order to evaluate our system’s validity. 

\begin{table}[h!]
\centering
\caption{The NEWS dataset labeled descriptive statistics.}
\label{table:news:dataset}
\begin{tabular}{cl}
\hline
Israeli perspective & 54 \\
Palestinian perspective & 59 \\
Neutral & 46 \\ \hline
\end{tabular}
\end{table}

We used two experienced human annotators who are native English speakers with advanced college degrees. We chose annotators that we felt displayed total integrity in their ability to follow the task assigned to them irrespective of their personal political opinion. We provided them with a list of the 159 articles and an instructional page with guidelines  which were very clear and limited in scope as seen in Figure \ref{fig:instructions}.  This enabled the annotators to perform the task while without injecting their personal political views. 

Before annotation of the articles, we advised our annotators to try to indicate which of the two sides wrote the article, i.e., whether the article seemed to be from an Israeli or Palestinian perspective.

\begin{figure}[h!]
    \centering
    \includegraphics[width=0.7\textwidth]{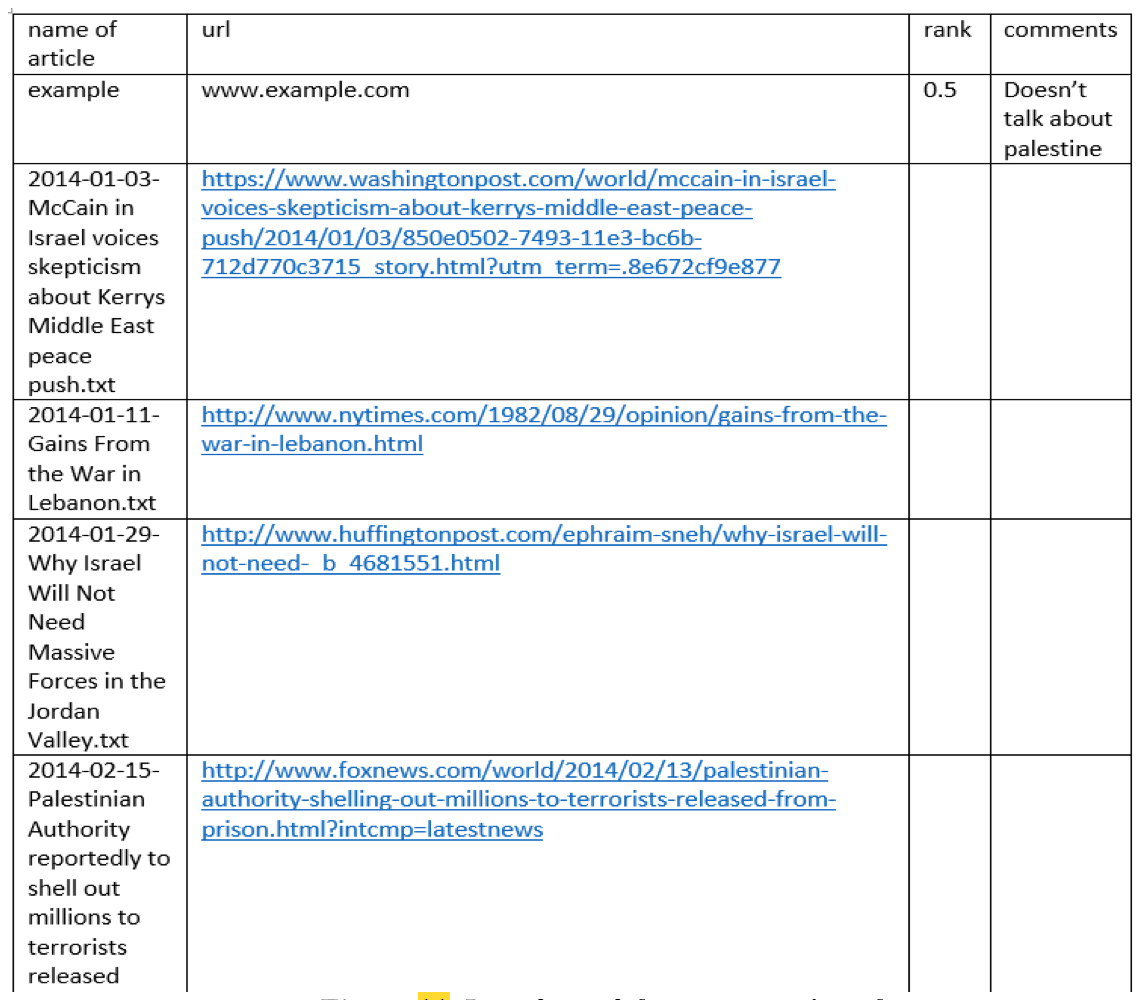}
    \caption{Annotator's interface} 
    \label{fig:interface}
\end{figure}

Figure \ref{fig:interface} presents the annotators' interface.

Then we compared the classifications of the human annotators. 
We conducted an inter-annotator reliability test with a high satisfactory agreement of 71\% ($kappa= 0.56$, $p < 0.05$) between annotators. The pair-wise kappa measure we obtained between annotators is considered to indicate a moderate level of agreement. The 112 articles that the annotators agreed upon served as our gold standard. 
We refer to this dataset as the NEWS dataset\footnote{ Available at \url{https://drive.google.com/open?id=1N2VVys1GDP2q\_jyPmh141p8f75lQq2jj} }.

\FloatBarrier
\subsection{CQA Dataset}
Our CQA dataset, used for the CQA task, includes questions from Yahoo Answers, a large and diverse CQA website, which has been active for over a decade
~\citep{adamic08ya}. 

The questions for our dataset were selected uniformly at random from the entire collection of \YA non-deleted English questions, posted in the years $2006$-$2016$. For our experiments, each question included its description ($81.5\%$ had non-empty descriptions), 
and list of answers with the answer's text and timestamp, in addition to its title and timestamp.
Each answer included a flag indicating whether it was selected as the ``best answer''~\citep{adamic08ya}. 

In our dataset questions are manually labeled as either `informational' or `conversational'. In principle, as defined by \harper, informational questions seek for facts or advice, while conversational questions pursue opinions, polling, or self-expression. As input for the labeling process, annotators were given the question's title, description, low level category, and top level category. In the first stage of the labeling process, two of the annotators went through an iterative process to form more specific guidelines, annotating three batches of $100$ questions each and carefully reviewing the disagreements. 
The level of agreement increased from $75\%$ for the first batch to $93\%$ for the third batch. The two authors then labeled a set of $1088$ additional questions with a level of agreement of $93.75\%$. For our experiments, we used the $1020$ questions for which there was an agreement between both annotators. 

To increase the size of our data, we asked students of the ``Introduction to Information Retrieval'' course to label additional questions from our random sample. Each student labeled a batch of $105$ questions. The batch included a benchmark of five questions for which the label was rather obvious. These five questions were randomly mixed together with 100 ``genuine'' questions. The students were graded based on their success in labeling the five benchmark questions\footnote{Students were unaware of the exact grading method.}
 (note that this grade accounted for a small portion ($2\%$) of the students' grade fro the course). The students received detailed written guidelines, based on the insights gained by the two authors during their annotation process, with definitions and examples for both types of questions. 
Each question was assigned to two students. Overall, $83$ students completed their annotations, with 76 ($91.6\%$) correctly answering all five benchmark questions. We discarded the input from the other seven students. The level of agreement between the student annotators was $84.7\%$, yielding a total of $2996$ additional questions whose label was agreed upon between the two annotators. Our complete dataset\footnote{http://webscope.sandbox.yahoo.com/catalog.php?datatype=l\&did=82} includes $4016$ labeled questions -- $1020$ from the annotators and $2996$ from the students -- of which $55.6\%$ are labeled informational and the rest are labeled conversational.

In addition to the labeled dataset, we created a large unlabeled dataset by sampling 20 million questions 
uniformly at random from the entire \YA archive of non-deleted English questions from $2006$ to $2016$. We refer to this dataset as the CQA dataset in this work .

\section{Evaluation}

In this section we describe the evaluation of the LSTM-based method on the two tasks of implicit dimensions identification. First, we use the same baseline algorithms for the two tasks. Then, we show how performance improved by using an LSTM network. Since each task we consider in the course of this research is a binary classification task we use the following performance evaluation metrics which are customary used in such cases \citep{Sokolova:2009:SAP:1542545.1542682}:
\begin{itemize}
\item Class-1 Recall. Measuring the ability of the classifier to correctly classify the Class-1  instances. 
\item Class-2 Recall. Measuring the ability of the classifier to correctly classify the Class-1  instances. 
\item AUC. Measuring the area under the ROC curve, which provides a general metric of the classifier's performance.
\end{itemize}
Class-1 recall and Class-2 recall have been referred to as specificity and sensitivity in previous research \citep{harper09facts} accordingly.
Other work \citep{Sarmento:2009:ACR:1651461.1651468} showed that the recall is a challenging measure for the task of UGC classification.
The baseline classification algorithms we employ in both tasks are SVM, Naive Bayes, Random Forest, Decision Tree and Logistic Regression. All of the baseline models were implemented and evaluated using the \texttt{sklearn} toolkit \citep{scikit-learn}.

We implemented\footnote{The CQA task models are available at: \url{https://github.com/vicmak/Sequence-classification}} \footnote{The NEWS task models are available at: \url{https://github.com/vicmak/News-Bias-Detection}} all of the LSTM models using Tensorflow~\citep{abadi16tensor} with the Keras API~\citep{chollet15keras}. We used the NLTK~\citep{bird06nltk} for text processing including sentence splitting and tokenization. We added a special artificial sentence's beginning and ending identification tokens.

\subsection{Political Perspective Identification}
\subsubsection{Baseline}
The results of the baseline algorithms' are presented in Table \ref{table:news:baseline}.  
We experimented with different features representing the articles during the baseline experiments. As in the case of CQA task we tried unigram, bi-gram and tri-gram features. We also identified that tf-idf features used as signal representation achieved a higher result. We used the following baseline algorithms and configurations:
\begin{itemize}
\item SVM with tf-idf features, using sigmoid kernel and penalty variable $C=1$.
\item Multinomial Naive Bayes classifier on tf-idf features with the alpha variable $C=1$.
\item Random Forest on tf-idf features with 10 estimators, splitting on the Gini criterion.
\item Decision Tree Classifier on tf-idf features. No limitation on max depth, splitting on the Gini criterion. 
\item Logistic Regression on tf-idf features using l2 penalty, with the inverse of regularization at 1 and maximum number of iterations 100. 
\end{itemize}

\begin{table}[h!]
\centering
\caption{Political perspective detection: Baseline classification results. }
\scriptsize
\label{table:news:baseline}
\begin{tabular}{lllll}
\hline
\textbf{Model} & \textbf{\begin{tabular}[c]{@{}l@{}}Classification \\ subject\end{tabular}} & \textbf{\begin{tabular}[c]{@{}l@{}}Class-1 \\ (Israeli) \\ Recall\end{tabular}} & \textbf{\begin{tabular}[c]{@{}l@{}}Class-2\\ (Palestinian)\\ Recall\end{tabular}} & \textbf{AUC} \\ \hline
SVM & Title & 0.926 & 0.397 & 0.796 \\
SVM & Content & 0.981 & 0.793 & 0.962 \\
SVM & Title + content & 0.833 & 0.586 & 0.710 \\
Naive Bayes & Title & 0.963 & 0.517 & 0.882 \\
Naive Bayes & Content & 0.963 & 0.741 & 0.901 \\
Naive Bayes & Title + content & 0.963 & 0.741 & 0.902 \\
Random Forest & Title & 0.833 & 0.621 & 0.799 \\
Random Forest & Content & 0.944 & 0.603 & 0.896 \\
Random Forest & Title + content & 0.870 & 0.483 & 0.795 \\
C 4.5 Decision tree & Title & 0.852 & 0.483 & 0.667 \\
C 4.5 Decision tree & Content & 0.889 & 0.707 & 0.798 \\
C 4.5 Decision tree & Title + content & 0.833 & 0.586 & 0.710 \\
Logistic regression & Title & 0.907 & 0.465 & 0.806 \\
\textbf{Logistic regression} & \textbf{Content} & \textbf{0.963} & \textbf{0.845} & \textbf{0.964} \\
Logistic regression & Title + content & 0.963 & 0.810 & 0.960
\end{tabular}
\end{table}
\FloatBarrier
The best baseline result was achieved using the logistic regression classifier, on the \textit{content} signal. Achieved Israeli perspective recall of 0.963, Palestinian perspective recall of 0.845 and AUC of 0.964. This result slightly outperforms the Title+content signal with the same logistic regression classifier. Another notable result is achieved by the SVM classifier on the content signal. It achieved a very high recall on the Israeli-perspective class, and a very high AUC value. 

\subsubsection{Improving the baseline - LSTM network}
 
\begin{table}[h!]
\centering
\caption{Political perspective detection: The results of LSTM-network classification. The configuration that lead to the best performance is marked with bold.}
\scriptsize
\label{table:lstm:news}
\begin{tabular}{llllllll}
\hline
\textbf{\begin{tabular}[c]{@{}l@{}}Embedding\\ type\end{tabular}} & \textbf{\begin{tabular}[c]{@{}l@{}}Classification\\ subject\end{tabular}} & \textbf{\begin{tabular}[c]{@{}l@{}}LSTM\\ size\end{tabular}} & \textbf{\begin{tabular}[c]{@{}l@{}}Word\\ cutoff\end{tabular}} & \textbf{\begin{tabular}[c]{@{}l@{}}Batch\\ size\end{tabular}} & \textbf{\begin{tabular}[c]{@{}l@{}}Class-1\\ (Israeli)\\ Recall\end{tabular}} & \textbf{\begin{tabular}[c]{@{}l@{}}Class-2 \\ (Palestinian)\\ Recall\end{tabular}} & \textbf{AUC} \\ \hline
Integrated & Title & 300 & 20 & 100 & 0.886 & 0.413 & 0.789 \\
Integrated & Title & 300 & 20 & 200 & 0.943 & 0.431 & 0.808 \\
Integrated & Title & 300 & 20 & 300 & 0.849 & 0.396 & 0.782 \\
Integrated & Content & 300 & 100 & 200 & 0.943 & 0.741 & 0.868 \\
Integrated & Content & 300 & 200 & 200 & 0.905 & 0.793 & 0.908 \\
Integrated & Content & 300 & 300 & 100 & 0.943 & 0.758 & 0.912 \\
Integrated & Content & 300 & 300 & 200 & 0.886 & 0.879 & 0.907 \\
Integrated & Content & 300 & 400 & 200 & 0.811 & 0.775 & 0.874 \\
Integrated & Title+content & 300 & 200 & 200 & 0.924 & 0.775 & 0.896 \\
GloVe & Title & 300 & 20 & 200 & 0.943 & 0.482 & 0.823 \\
GloVe & Content & 300 & 300 & 200 & 0.924 & 0.655 & 0.903 \\
GloVe & Title+content & 300 & 300 & 200 & 0.830 & 0.741 & 0.873 \\
Newswire & Title & 300 & 20 & 200 & 0.905 & 0.327 & 0.779 \\
Newswire & Content & 300 & 300 & 200 & 0.867 & 0.793 & 0.915 \\
Newswire & Title+content & 300 & 300 & 200 & 0.905 & 0.827 & 0.932 \\
Newswire & Title+content & 600 & 600 & 50 & 0.911 & 0.810 & 0.911 \\
Newswire & Title+content & 600 & 600 & 100 & 0.962 & 0.879 & 0.953 \\
Newswire & Title+content & 600 & 600 & 150 & 0.962 & 0.793 & 0.953 \\
Newswire & Title+content & 600 & 600 & 200 & 0.824 & 0.758 & 0.939 \\
\textbf{Newswire} & \textbf{Title+content} & \textbf{800} & \textbf{800} & \textbf{100} & \textbf{0.966} & \textbf{0.879} & \textbf{0.966} \\
Newswire & Title+content & 900 & 900 & 100 & 0.905 & 0.810 & 0.927
\end{tabular}
\end{table}
\FloatBarrier
We intended on improving the baseline by using LSTM for two reasons. First, employ the LSTM model for capturing long term linguistic regularities, which might be harder to capture by simpler classification models which use n-gram or tf-idf features. Second, the ability to use an unlabeled data for pre-trained word embeddings.
Table \ref{table:lstm:news} presents the results of the LSTM network classification. We implemented our classifier as described in section 3 (Method Section) and performed multiple experiments with the following settings and hyper-parameters:

\begin{itemize}
\item All possible combinations of article's title, its content, and the combination of title+content - a concatenation of the two pieces of text.
\item The source of the word embedding. First, we experimented with integrated embedding, which is learned alongside all other network's parameters. Second, we train our proprietary embeddings, using only the domain-specific text of all the news text we collected as described in the news-dataset section. Third, to explore pre-trained word embeddings further, we experimented with pre-trained\footnote{Available at \url{https://nlp.stanford.edu/projects/glove/}} GloVe \citep{pennington2014glove} embeddings that were trained on general purpose text from Wikipedia and Gigaword.  
\item The LSTM memory size. We experiment with the size of the LSTM's cell $c_t$, which is computed (equation 10) during each and every LSTM time stamp during its training and prediction. The LSTM's size is responsible for the robustness of its memory, its ability to remember both long and short history.
\item Word cutoff. We experimented with the size of the signal which is used for the classification. We set different lengths of the text, starting from the article's beginning which was fed into the LSTM network. Put otherwise, we experimented with various wide depth of the network, during its \textit{unrolling}.
\item Batch size. We experimented with the batch size during the training phase. It is very impractical to train such a deep LSTM network with single stochastic training. However, the batch size affects the convergence process \citep{Goodfellow-et-al-2016} and the final classification results. We experimented with a cutoff of 20 at the title signal only, and several cutoffs of hundreds at the title+content, and content signals.
\end{itemize}

The best result with an enhanced performance is achieved when we train a relatively large LSTM network, with 800 words cutoff, a $c_t$ memory cell size of 800 and on a Title+content signal. This configuration outperforms the best baseline configuration of logistic regression in all three metrics, and achieves notably high recall in the Israeli-perspective class and in the AUC, both of value 0.966. This performance was achieved not only because of a very deep and complex network structure which was a result of a long series of experiments, but also because of the usage of the pre-trained word embeddings, that were trained with the SGNS \citep{mikolov13efficient} mode by the gensim \citep{rehurek10software} toolkit. As in all other models in the course of this research, the pre-trained embeddings were used only as weights initialization, and were further updated during the training process. 

\begin{table}[]
\centering
\caption{International news articles classification outputs statistics. The Israeli label in this binary classification task was 1, and the Palestinian 0. In the table below under 0.5 means leaning to the Palestinian perspective, and above 0.5 means leaning to the Israeli perspective.}
\label{table:american:british}
\begin{tabular}{lcccc}
\hline
 & BBC & Fox News & New York Times & The Guardian \\ \hline
Number of articles & 536 & 1299 & 1980 & 260 \\
Average output & 0.260 & 0.563 & 0.491 & 0.449 \\
Median output & 0.143 & 0.628 & 0.500 & 0.398 \\
Standard deviation & 0.276 & 0.358 & 0.332 & 0.332 \\
Articles under 0.5 & 428 & 550 & 984 & 152 \\
Articles over 0.5 & 104 & 743 & 980 & 113 \\
Percentage under 0.5 & 0.798 & 0.423 & 0.496 & 0.584 \\
Percentage over 0.5 & 0.201 & 0.576 & 0.503 & 0.415 \\ \hline
\end{tabular}
\end{table}
\subsubsection{British vs. American Press}
In the performance test phase, when we used the articles which were labeled by human annotators from an  international press. In the training phase we used the Israeli and the Arab press to represent the leaning perspective of each class.
We decided to perform a wider examination on all international articles, and see whether there are some leanings to either of the sides, as our model may suggest. We use the best performing LSTM-network model to classify all international articles we gathered and look at the output, given its probabilistic interpretation of belonging to one of the classes.

Table \ref{table:american:british} presents the statistics of the LSTM classifier's output on the international press articles.
According to our model, there are differences in perspective among four international online media outlets: Fox news, The New York Times, The Guardian and BBC.  Our findings clearly show that British news perspective is leaning more towards the Palestinians while American news sources are more neutral and even slightly tilted towards Israel.  

\subsection{Identifying Informational vs Conversational Question in CQA Archives}

\subsubsection{Baseline}
Our baseline replicates the text-based classifier described by \citep{harper09facts} over our CQA dataset. We used the Weka workbench~\citep{hall09weka} to run our classifier and five-fold cross validation to evaluate performance. We use the same three metrics as \citep{harper09facts} for measuring the classifiers' performance: (i) Class-1 Recall (sensitivity) -- the proportion of conversational questions that are correctly classified; (ii) Class-2 Recall (specificity) -- the proportion of informational questions that are correctly classified; and (iii) AUC - area under the ROC curve.

The text classifier used by \citep{harper09facts} was based on the $500$ most common unigrams or bigrams occurring in the titles of each of the two types of questions (in lower-case form). For classification, Weka's sequential minimum optimization (SMO) algorithm for training support vector machines (SVM), with linear kernel, was used. Performance results can be seen in the first section of Table~\ref{table:rescqa}.

\begin{table}[]
\caption{Performance results for the CQA task using different methods. For the LSTM-network classifier we also specify the standard deviation of the folds.}
\label{table:rescqa}
\resizebox{\textwidth}{!}{%
\begin{tabular}{lllll}
\hline
Model & Classification subject & Class-1 Recall & Class-2 Recall & AUC \\ \hline
\citep{harper09facts} & Title & 0.667 & 0.845 & 0.756 \\ \hline
LSTM & Title & 0.832+-0.051 & 0.794+-0.043 & 0.899+-0.009 \\
LSTM & Title+description & 0.874+-0.043 & 0.774+-0.100 & 0.909+-0.011 \\
LSTM & Title+description+best answer & 0.788+-0.170 & 0.831+-0.064 & 0.897+-0.021 \\ \hline
\end{tabular}%
}
\end{table}

\subsubsection{Improving the Baseline - LSTM Network}

To improve the baseline proposed by \citep{harper09facts}, we used RNNs with LSTM. In our experiments, we lower-cased and tokenized the text and fed it sequentially (from left to right) into the model. 
We experimented with different textual fields as input to our classifier: title-only, title+description, title+description+answers, and title+description+best answer. Among the LSTM models we tested, we achieved the highest performance in the model that used concatenated title, description and best answer as a classification signal. This model outperformed the baseline in Class-1 Recall and AUC. It was also slightly less by 0.014 in the Class-2 Recall metric.


In order to obtain statistically valid results in our cross-validation experiments \citep{Salzberg97oncomparing, varma2006bias} we did not perform any further hyper-parameter tuning. We used the hyper-parameters' values from the NEWS task experiment, that include:
\begin{itemize}
\item The cutoff of 30 words for title, cutoff of 200 words for title and description, and a cutoff of 800 words for the combination of title, description and best answer.
\item Batch size of 100 instances.
\item LSTM memory $c_t$ size equal to the cutoff length, that is 30,200 and 800 accordingly.
\item A default no-dropout use of the Keras library.
\end{itemize}

We pre-trained skip-gram negative sampling (SGNS) \citep{mikolov13efficient} word embeddings over our unlabeled dataset using the Gensim library~\citep{rehurek10software} and experimented with three variants of the text used for training: titles only; titles and descriptions; and titles, descriptions, and  answers. In all cases, performance differences among the three were minor, with the title-only text variant achieving slightly higher performance than the others. Therefore, we only report title only embeddings, which also require less storage and computation power.


\subsubsection{Model Architecture Change}
We also experimented with a bi-directional RNN with LSTM, however this model did not yield any performance gain. Our model's code is available via a public git repository\footnote{https://github.com/vicmak/Sequence-classification}.

\subsection{Hyper-Parameters}
Deep Neural Networks are quite hard to train, and there is a need of huge data resources, to achieve an enhanced performance. The LSTM network is not an exception. It has complex structure which is hard to completely understand, not easy to train, and it tends to overfit unless a huge amount of data is available at the training time.
\begin{enumerate}
\item LSTM size $c_t$. Is the size of the LSTM memory cell. The bigger LSTM cell's size is, the more robust is its ability to learn long-term dependencies. On the other hand - setting a this hyper-parameter relatively big leads to a more complex model, with more parameters to be learned, and increasing the training time - making the overall experiment process much harder. In the NEWS task the best result was achieved with the cell size set to 800. In the CQA task the best result was achieved with the cell size set to 200.  We find this result reasonable, since in the CQA task the word cutoff used for classification is smaller, and thus the need for long linguistic patterns remembering.
\item Embedding size. Embedding size can be varied due to different vocabulary sizes. In our case, as in other previously reported work \citep{melamud2016context2vec, pennington2014glove} that included embedding size tuning, the dimensionality of 300 achieved a superior performance. While it is possible to train embeddings with even larger dimensions, the performance improvement is negligible and the training time of the model is significantly bigger, making this larger dimension impractical and not cost effective.
\item Integrated vs Proprietary embedding source. The enhanced performance that is usually achieved by deep LSTM networks owes it to the availability of large data in the training phase. In the case of text classification, the immediate suspect for the large training data is the ability to use a pre-trained embeddings. Note that embedding is a significant portion of the model's parameters, since they are actually represented as a matrix which transforms the one-hot vectors of vocabulary size into a lower space. We show in our experiments the improvement achieved by setting the initial value of these embedding parameters on large unlabeled data. We use the proprietary domain specific data for each task, and probably the most prominent \textit{word2vec} \citep{mikolov13efficient} toolkit with the SGNS embedding type. While there are several word embedding techniques that were introduced over the years, it was recently shown \citep{LevyEmbedding} that the different techniques vary by their implementation design and hyper parameters choice based on a specific task.
\item Batch size is known to have an effect \citep{Goodfellow-et-al-2016} on the network's convergence, and thus on the final model's classification performance. Since a stochastic training is virtually impossible due to time limitations we performed a fine tuning of the batch size by multiple experiments. We find that the task of batch size selection should be treated very carefully, and multiple experiments should be run before achieving the most enhanced performance. In our experiments the optimal sizes we between 100 and 200 instances of text pieces for training. In the Informational vs. Conversational questions identification the most effective batch was of size 200. In the political perspective the most effective batch size was 100. In both tasks batches smaller than 100 and larger than 200 lead to classification performance decrease. 

\item Bidirectional LSTM architecture.

\begin{figure}[h!]
    \centering
    \includegraphics[width=0.5\textwidth]{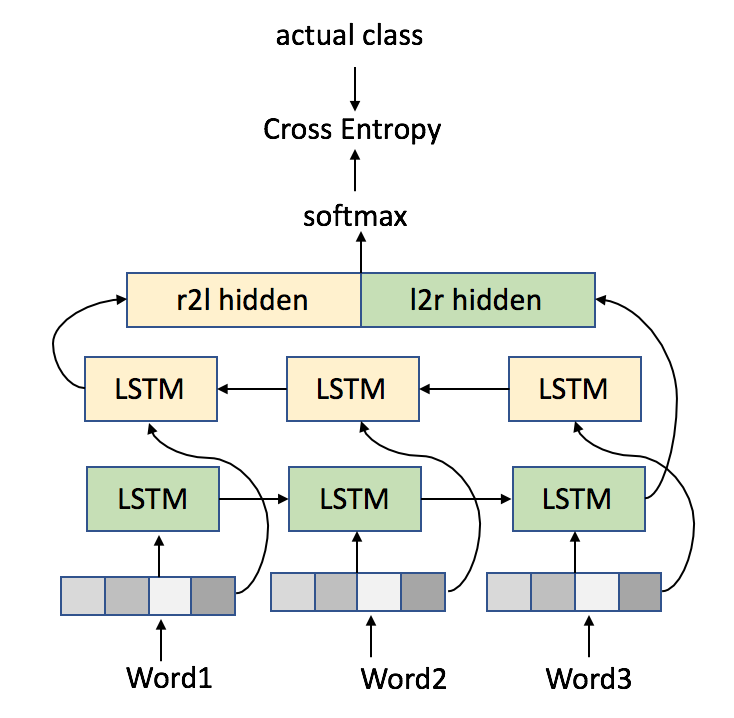}
    \caption{The bidirectional LSTM network architecture for a binary classification task. } 
    \label{fig:bilstm}
\end{figure}
\FloatBarrier
We also experimented with changing our architecture of the LSTM network, as shown in Figure \ref{fig:bilstm}. We tried a bidirectional LSTM architecture \citep{2015arXiv151100215W,P16-2067,melamud2016context2vec}, which is actually a concatenation of two two different LSTM networks which are trained simultaneously, and then classified as a single, but more complicated representation. We sought to check whether the order of the words in a long-term linguistic regularity might affect the classification quality. In both the CQA and the NEWS task we failed to outperform the single direction LSTM model performance with a bidirectional LSTM model. 
In the CQA task the best bidirectional performance achieved was \textit{Class-1 Recall = 0.756}, \textit{Class-2 Recall = 0.804} and \textit{AUC = 0.872}. In the NEWS task the best  bidirectional performance achieved was \textit{Class-1 Recall = 0.924}, \textit{Class-2 Recall = 0.655} and \textit{AUC = 0.893}.
\end{enumerate}

\section{Conclusions}

We introduced two interesting and novel datasets for the task of implicit dimensions identification. We investigated the two datasets through appropriate tasks: (1) the NEWS task - political perspective identification in online news articles, and (2) the CQA task - Informational vs. Conversational Question identification in CQA archives.

In the NEWS task we demonstrated how local press articles, Israeli and Arab, can be used to train classifiers that reflect the author's perspective. We demonstrated our results on held out set of articles from presumably neutral media outlets from the UK and the USA. 

 In both tasks we performed a vast amount of experiments with various classification algorithms, to explore how the implicit dimension identification task, can be modeled as a text classification problem.

In the NEWS task, we demonstrated the LSTM network's capabilities that outperform the baseline methods. We have shown that an LSTM network, a statistical classifier with a complex nonlinear structure and multiple parameters that need to be learned, outperforms the more standard and used classifiers - the Naive Bayes, Decision Tree, Logistic Regression, SVM and Random Forest. 
However, We see that the LSTM network is not a silver bullet for this text classification problem. For example, in Table \ref{table:lstm:news} - most of the configurations of the network itself, combined with various signals fail to outperform the best baseline result achieved by the folklore logistic regression classifier.  

Moreover, in the CQA task the LSTM-network outperformed the baseline only in the AUC metric in all combinations of classification subject.

We showed how the unlabeled data can assist the classification task, using it for proprietary embedding pre-training. In the CQA and NEWS tasks, we used these pre-trained embeddings to initialize the networks input weights, and updated them to achieve the most enhanced performance.

For the future work we intend on exploring more methods to tackle the common machine learning scenario when there is a high availability of unlabeled data aside a relative small amount of labeled data, such as label propagation and the Ladder Network \citep{DBLP:conf/nips/Premont-Schwarz17,DBLP:conf/nips/RasmusBHVR15}.

\bibliography{example,infconv}

\end{document}